\documentclass[sigconf]{acmart}
\AtBeginDocument{%
  \providecommand\BibTeX{{%
    \normalfont B\kern-0.5em{\scshape i\kern-0.25em b}\kern-0.8em\TeX}}}
\acmDOI{10.1145/3520304.3528992} %
\acmConference[GECCO '22]{The Genetic and Evolutionary Computation Conference 2022}{July 9--13, 2022}{Boston, USA}
\acmYear{2022}
\acmISBN{978-1-4503-392-686}

\author{M. Zameshina$^{1,2}$, O. Teytaud$^{2}$}
\affiliation{
\institution{1. LIGM, Univ Gustave Eiffel, CNRS, ESIEE, Paris\\2. Facebook AI Research}
\country{France}
}
\email{{mzameshina,oteytaud}@fb.com}

\author{Fabien Teytaud}
\affiliation{
\institution{Univ. du Littoral Cote d'Opale}
\country{France}
}
\email{teytaud@univ-littoral.fr}

\author{Vlad Hosu}
\affiliation{
\institution{Univ. of Konstanz}
\country{Germany}
}
\email{vlad.hosu@uni-konstanz.de}

\author{Nathanael Carraz}
\affiliation{
\institution{Univ. d'Antananarivo}
\country{Madagascar}
}
\email{carraznathanael@live.fr}

\author{Laurent Najman}
\affiliation{
\institution{LIGM, Univ Gustave Eiffel, CNRS, ESIEE Paris}
\country{France}
}
\email{laurent.najman@esiee.fr}

\author{Markus Wagner}
\affiliation{
\institution{The University of Adelaide}
\country{Australia}
}
\email{markus.wagner@adelaide.edu.au}

\usepackage[utf8]{inputenc}
\usepackage{placeins}

\title{Fairness in Generative Modeling: do it Unsupervised!}
\usepackage{multicol,multirow}
\usepackage{algpseudocode,algorithm}

\usepackage{graphicx,url}
\def\R{\mathbb{R}}

\begin{document}

\begin{abstract}
We design general-purpose algorithms for addressing fairness issues and mode collapse in generative modeling.
More precisely, to design fair algorithms for as many sensitive variables as possible, including variables we might not be aware of, we assume no prior knowledge of sensitive variables: our algorithms use unsupervised fairness only, meaning no information related to the sensitive variables is used for our fairness-improving methods.
All images of faces (even generated ones) have been removed to mitigate legal risks.
\end{abstract}

\maketitle

\section{Introduction}
Fairness has become prevalent at the intersection of ethics and artificial intelligence. 
Various forms of fairness are critical in online media~\cite{fairnessonline}.
In the present paper, we consider fairness in the context of generative modeling. 
More precisely, when modeling the probability distribution of faces, we typically observe that classes already rare in the dataset become even rarer in the model. This phenomenon is called Mode Collapse (MC)~\cite{defmc}, and for sensitive variables, it is one of the fairness issues.
We propose tools based on statistical reweighting (Sections~\ref{troisun} and~\ref{troisdeux}) or on user feedback (Section~\ref{troistrois}) for mitigating  fairness issues (such as MC) in generative modeling.

\subsection{Fairness} 
There are many facets to fairness. An algorithm may be considered to be fair if its results are independent of some variables, particularly for sensitive variables.
Fairness~\cite{fairness} can be measured in terms of separation, i.e., whether the probability of a given prediction, given the actual value, is the same for all values of a sensitive variable. The measurement can also be rephrased in terms of equivalent false negative and true negative rates for all classes.
A distinct point of view is sufficiency: sufficiency holds if the probability of actually belonging to a given group is the same for individuals from that group and with different sensitive variables. 
Another point of view is independence, i.e., when the prediction is statistically independent of sensitive variables.
Because it is known that the many criteria for fairness are contradictory, it is necessary to design criteria depending on the application. In the present paper, we consider the case in which the goal is to preserve some frequencies.

Here, we consider the context of generative modeling. There is a model trained on data, and we want this model to satisfy some requirements on frequencies: for every class, we would like the frequency to match some target frequency. Typically, for simplicity in the present paper, the target frequency is the frequency in the original dataset: however, the methods that we propose can be adapted to other settings.

\subsection{Generative modeling: fairness and mode collapse}\label{mcrare}
There are many measures of fairness, even in the specific case of generative modeling~\cite{measuringfairnessgm}. 
The main criterion is whether all classes are correctly represented. It is known that modeling frequently decreases the frequency of rare classes (i.e., mode collapse). In addition, improving the image quality (for each image independently) aggravates the diversity loss~\cite{salminen}. For a conditional generative model, there is sometimes a ground truth. For example, in super-resolution, we want the reconstructed image to match the sensitive variables of the ground truth as closely as possible. This case became particularly critical since, e.g., \cite{srganbias}: a pixelized version of Barak Obama can be ``depixelized'' to be that of a white man.
\cite{latersup} points out the importance of fairness in the design of Generative Adversarial Networks (GANs) before applying them, for example as an early stage before supervised training.
For addressing fairness issues, a possibility is to increase editability:
\cite{fairgan} disentangles latent variables for separating editable and sensitive parts. Some works focus on measuring fairness, for example,  \cite{faircounter} uses causal methodologies for measuring fairness in a counterfactual manner. Fairness can be integrated directly into the training: \cite{fairnessgan} focuses on training a GAN while protecting some variables.

\subsection{Related work}
\cite{tirloulou} increases fairness in GANs in a supervised manner, i.e., given the sensitive attributes. \cite{otherfairgan} targets and improves the fairness of generated datasets. More similar to our work, \cite{uncertain} focuses on uncertain sensitive variables, and \cite{biasgan} adds a bias in a GAN for mitigating fairness issues. In the same fashion as the present work, \cite{fairgen} considers biasing a GAN without any retraining.
We focus on generically (i.e., independently of the application, data, and model) correcting for potential bias present in a generative model, {\em{without knowing the sensitive variables}}. The critical point is that sensitive variables seem to often come up as a surprise: typically, people do not decide to create an unfair algorithm actively. For example, in~\cite{watergate}, the designers of the faulty soap dispenser had just not imagined that it might fail on black skins. Also, there may be relevant sensitive variables that have not been initially considered: ethnicity or gender are obvious sensitive variables, but aesthetics, body mass index, social origin, or even the quality of the camera, geographical origin, also matter. 

Our goal is to have a generic correction independent of the sensitive variables. The first proposed method (Sections~\ref{troisun} and~\ref{troisdeux}):
\begin{itemize}
	\item is not only for the fairness issues regarding sensitive variables: we also preserve diversity for more classical diversity issues such as MC. %
	\item does not need any retraining.
	\item is more or less effective depending on cases but is designed for (almost) never being detrimental (Section \ref{cool}).
\end{itemize}
The second proposed method, which can be combined with the previous one, proposes several generations and then lets the user choose. Therefore, the user experience is modified: we expect the user to assist the method by actively selecting relevant outputs. Contrary to the generic method proposed above, which we will implement thanks to reweighting, the new approach is not a drop-in replacement. Moreover, this also does not need retraining.

\subsection{Outline}
Section~\ref{deux} presents tools useful for the present work:
\begin{itemize}
	\item Use of Image Quality Assessment (IQA) to improve image generation (Section~\ref{deuxun}): we connect this method to our research by investigating how much this quality improvement degrades fairness and how our proposed methods can mitigate such issues.
	\item Reweighting via simple rejection sampling to improve fairness and reduce MC when the variables used for computing the reweighting values are correlated to the target sensitive variables (Section~\ref{troisun}).
\end{itemize}
Section~\ref{trois} presents our proposed algorithms:
\begin{itemize}
	\item Reweighting as above, but with reweighed variables unrelated to target classes (Section~\ref{troisdeux}). This second context is therefore applicable when we do not know the target classes. We propose a method which is a drop-in improvement of an arbitrary generative model: as soon as we have features and a generative model, we can apply Alg.~\ref{algrw}.
    \item Multi-objective optimization, through computation of several solutions (typically Pareto fronts), to mitigate diversity loss by providing more frequently at least one output of the category desired/expected by the user.
\end{itemize}
Section~\ref{quatre} is a mathematical analysis.
Section~\ref{cinq} presents experimental results.

\section{Preliminaries}\label{deux}
\subsection{Correlations image quality / sensitive variables}\label{deuxun}
We investigate the known correlation between the estimated quality of an image and its membership to a frequent class~\cite{salminen,pulse}.

In order to demonstrate that this is easily observable, Table~\ref{zecor} presents the rank correlation between the aesthetic quality of an image and the logit of that image for each of four classes of individuals. We note that the most positively correlated class is the most frequent. %
Our interpretation is that the technical quality of generated images is higher for the most frequent classes, influencing the aesthetics score. %

\begin{table}[t]\footnotesize \centering\centering
\begin{tabular}{|c|cccc|}
\hline
Class & A & B & C & D \\
\hline
	Frequency & 17.8\% & {\bf{52.2\%}} & 17.5\% & 12.4\% \\
	Rank-correlation AvA & -0.07 & {\bf{0.22}} & -0.11 & 0.06  \\

	Rank-correlation K512 & -0.02 & {\bf{0.16}} & -0.08 & 0.02 \\
\hline
\end{tabular}
	\caption{\label{zecor}For four distinct classes of individuals A, B, C and D (obtained using R), we present the rank-correlation of the frequency of that class with AvA and K512 scores respectively. AvA and K512 are visual quality estimators, dealing with aesthetics and technical quality respectively. Visual quality assessment is a task fairly independent of semantics and therefore should exhibit little if any ethnicity-related biases. \label{tabun} Dataset: faces generated by StyleGan2 (see \url{thispersondoesnotexist.com}). Classes: ethnicity evaluated by R (see R in Table~\ref{tab:fe}). Observation: the biggest class has the strongest, positive correlation.}
\end{table}

\subsection{Image generation: GAN, PGAN, and EvolGan}\label{deuxdeux}
Our work specializes in image generation, and in particular on faces. We use the following image generation tools. Our baseline GAN is Pytorch GAN Zoo (\cite{pytorchganzoo}, based on progressive GANs (PGANs)~\cite{karras2017progressive}). We also use EvolGan~\cite{roziere2020evolgan}, which improves Pytorch GAN Zoo by biasing the random choice of latent variables $z$ using K512~\cite{koncept512reference}. We use three configurations of EvolGan, as it uses as a budget the number of calls to the original GAN; the three configurations then correspond to budgets 10, 20, and 40 (named $EG10$, $EG20$, and $EG40$ respectively).
Besides the one based on a random search, EvolGan has an option for CMA search~\cite{HAN} and PortfolioDiscrete-$(1+1)$ (i.e. the variant of the Discrete $(1+1)$-ES as in~\cite{danglehre}): we also employ these variants, with notation respectively EG-CMA-10 and EG-D$(1+1)$-10 for budget 10, and similar variants for budget 20 and 40. Therefore we have nine flavors of EvolGan, corresponding to different algorithms and budgets.

\subsection{Diversity loss in generative modeling}\label{genmod}
Usually, modeling decreases the frequency of rare classes.
With StyleGan2, we get 71.55\% white people and 4.64\% black according to R (close to~\cite{salminen}). EvolGan, which is built on top of StyleGan2 with a budget of 40 decreases the percentage of black people to 0\% while increasing the frequency of white to 81.25\%.

\subsection{Measuring the diversity loss}\label{dl}
We assume that there exist target frequencies for each sensitive class. In the present paper, we focus on preserving the diversity in the sense of ``having the same frequencies as the frequencies in the original data used for creating the model'', so the target frequencies are the frequencies in the original dataset. If we consider the diversity loss associated with optimizing a model, such as EvolGan, we assume that target frequencies are those of the original model.

Given classes $\{1,\dots,n\}$ with target frequencies $f_i$ ($\sum_{i=1}^n f_i=1$), and real frequencies $f'_1,\dots,f'_n$: the diversity loss $\Delta$ is defined as 
$\Delta:=1-\inf_{f_i>0} f'_i/f_i$. $\Delta=0$ if the target frequencies are reached, and $\Delta=1$ if one of the classes has disappeared.
Throughout our paper, we consider diversity loss for classes, and not inside each class: this other important case is left as further work.

\subsection{Feature extractors}\label{secfe}
We use various feature extractors (Table~\ref{tab:fe}). E and R use VGG-Face~\cite{vggface}. The goal of these feature extractors is to have auxiliary classes for reweighting: these values, after discretization, provide classes. These classes, termed strata, are used in Section~\ref{troisdeux}.

\begin{table}[t]\centering
    \centering\footnotesize
    \begin{tabular}{|cccc|}
    \hline
     Name    & Notation & Domain & Note \\
     \hline
     \multicolumn{4}{|c|}{Variables to be protected}\\
    \hline
    R     & $R$ & $\{A,B,C,D\}$ & Ethnicity \cite{re}\\
    AvA     & $AvA$ & $\{F,E\}$ & Aesthetics\cite{ava} \\
    \hline
    \multicolumn{4}{|c|}{Related auxiliary variables}\\
    \hline
               & $R'$    & $\R^4$ & Logits of $R$\\
    Koncept512 & $K512$  & $\R$   & IQA          \\
    \hline
    \multicolumn{4}{|c|}{Unrelated auxiliary variables}\\
    \hline
    Emotions   & $E$     & $\{1,2,3,4,5,6,7\}$         & facial expression in  \cite{re} \\
               & $E'$     & $\R^{100}$ & final layer of $E$ \\
    \hline
    VGG-Face   & $VF$    & $\{0,1\}^{128}$ & Binarized \\
    final layer   &         &                 & VGG-face \\
    \hline
    \end{tabular}
	\caption{Feature extractors used in the present article. All data are faces, typically generated by StyleGAN2 or other methods in Section~\ref{deuxdeux}.\label{tabdeux}}
    \label{tab:fe}\vspace{-3mm}
\end{table}

\section{Methods}\label{trois}
Section~\ref{rw} presents a simple rejection method for ensuring target probabilities in generative modeling.
Section~\ref{rw2} shows how to build classes in order to apply that method without knowing what the sensitive variables are. Section~\ref{methodmoo} then presents a methodology based on multi-objective optimization for improving fairness. 

\subsection{Reweighting: stratified rejection}\label{rw}\label{troisun}
Consider a generative model on some domain $D$.
Consider a partition $D_1,\dots,D_m$ of $D$  into $m$ disjoint strata. Assume that some unknown random variable $\omega$ has probability $p_i=P(\omega\in D_i)$ and $\sum p_i=1$. We have another random variable $g$ also living with probability one in the union of the $D_i$.
Assuming that $P(g\in D_i)=p'_i$, a simple tool for building $g'$ such that $P(g'\in D_i)=p_i$ is rejection (see Alg.~\ref{algrw}). This simple algorithm generates $g'\in D_i$ with probability $p_i$.

\begin{algorithm}\vspace{-1mm}
\begin{algorithmic}
	\State Generate $x$ a (new, independent) output of $g$ \hfill $//$ random gen
	\State Find $i$ such that $x\in D_i$.
	\State With probability $1-\frac1{\max_j \frac{p_j}{p'_j}}\frac{p_i}{p'_i}$, go back to random gen.
	\State \Return $g':=x$.
\end{algorithmic}
	\caption{\label{algrw}Given a generative model $g$, bins $D_1,\dots,D_m$ and their target probabilities $p_1,\dots,p_m$. This algorithm assumes that none of the $D_i$ has probability $0$ for the original generative model~$g$.}\vspace{-1mm}
\end{algorithm}

\subsection{Creating strata: reweighting without knowing the target classes}\label{rw2}\label{troisdeux}
We have classes corresponding to sensitive classes.
We consider four sensitive classes of faces (A, B, C, D) using R~\cite{re} and two classes using AvA~\cite{ava} (class F = bottom 20\% of the aesthetics variable). 
However, we also want (possibly non-sensitive) classes used as auxiliary classes for reweighting: our goal is for our method to work for unknown target classes, so we need auxiliary classes. The idea is to investigate how much we can improve fairness for variables A, B, C, D without using those classes in our algorithm. Our auxiliary classes (Section \ref{secfe}), unrelated to our sensitive classes, will be called strata in the present work: the strata are the $D_i$ used in our reweighting algorithms.

The key point in our experiments ``preserving the diversity of unknown target variables'' is that we do not use the target variables in our algorithms: our method is unsupervised in this sense. %
When we try to maintain diversity for class F, we can use auxiliary variables which are unrelated to F: so, we can use A, B, C and D. And when we try to maintain diversity for classes A, B, C and D, we can use F as an auxiliary variable.

Some attributes (final layer %
of an emotion classifier, or technical quality of the photo) can be used for all classes as they are not directly related to any of our sensitive variables. 
We will use two parameters $d$ and $M$ in our experiments.
Given a possibly large number of auxiliary variables (not the target variables), we select $d$ variables.
Each of these $d$ variables is discretized in $M$ values, where $M$ is called the arity: thresholds are chosen so that the $M$ values are equally frequent.

\subsection{The user-assisted context: generating multiple solutions}\label{methodmoo}\label{troistrois}
Whereas in Section~\ref{rw2} we have considered a drop-in replacement of the baseline, which generates one image per instance, we now consider the case in which we generated several instances, and the user can select one of them (see Alg.~\ref{algmultiple}). 
There are two parts: how to generate multiple contexts, and, for some methods which generate way too many solutions for being manually searched by a human user, how to sample the obtained Pareto front.

\subsubsection{How to generate multiple solutions}
\begin{algorithm*}
	\begin{multicols}{3}
		{\bf{No context, no user assistance}}
		\begin{algorithmic}
				\State Repeatedly, generate one individual per request.
				\State Check that their frequencies match the expectation: compute a DL.
		\end{algorithmic}
		\columnbreak

		{\bf{Context, no user assistance}}
		\begin{algorithmic}
				\State Repeatedly, generate one individual per request. Requests have a context (e.g., low-resolution image).
				\State Check that their frequencies match the frequencies of the context (e.g., same ethnicity as low-res image): compute a DL.
		\end{algorithmic}\columnbreak
		{\bf{Context, user assistance}}
		\begin{algorithmic}
				\State Repeatedly, generate $k$ individuals per request (e.g., by Pareto-based MOO, or by diversity-based MOO, or by MSR): the user chooses one of them.
				\State Check that their frequencies match the contextual expectations: compute DL.
		\end{algorithmic}
	\end{multicols}
	\caption{\label{algmultiple}Different contexts for image generation, without or with human assistance. Left: unassisted context, generative model. Middle: generative model with target class (case in which there is an expected class, e.g., super-resolution in which the ethnicity is supposed to be preserved statistically). Right: user-assisted method. Not all unsupervised fairness methods can be applied in all cases. The reweighting method in Sections~\ref{troisun} and~\ref{troisdeux} can be applied to the two first columns. In contrast, the multiple generation such as the one in Section~\ref{troistrois} can be applied to the third column only.}
\end{algorithm*}

We consider a fixed limit on the number of generated images allowed so that the tool remains manageable for the user.
Several approaches can generate a targeted number of outputs; we consider (i) multi-objective optimization (MOO: splitting the original criterion into several and optimizing them jointly) and (ii) multiple runs.
Doing multiple runs is a simple and intuitive solution for generating multiple images.
Regarding MOO, our solution is not compatible with all generative models: we consider that images are obtained by numerical optimization of a linear combination of criteria~\cite{camilleinspir}. Instead of aggregating them, \cite{moocv}~proposed to preserve diversity by optimizing several numerical criteria by MOO, and we include this technique (as well as the previously mentioned reweighting techniques) in our fairness context. MOO naturally generates several solutions instead of one so that we are (presumably) more likely to have at least one satisfactory solution. 

\subsubsection{How to sample the obtained solutions}
When we do multiple runs, we can choose their number to control the number of generated images. However, in MOO, we typically get a Pareto front. This Pareto front might be huge. Therefore, we have to sample this Pareto front. There are many tools for this:
\begin{itemize}
    \item Optimizing this sampling for some representativeness criterion in the fitness space (hypervolume and others, see Appendix~\ref{subpareto}).
    \item Or maximizing some diversity criterion in the original domain, regardless of fitness values.
\end{itemize}%

\section{Methods analysis}\label{quatre}

\subsection{Multi-objective diversification}\label{moomaths}
Generating several solutions and letting the user choose among those proposals is a simple workaround for partially mitigating diversity loss.

\begin{figure}\centering\vspace{2mm}
	\includegraphics[width=.48\textwidth]{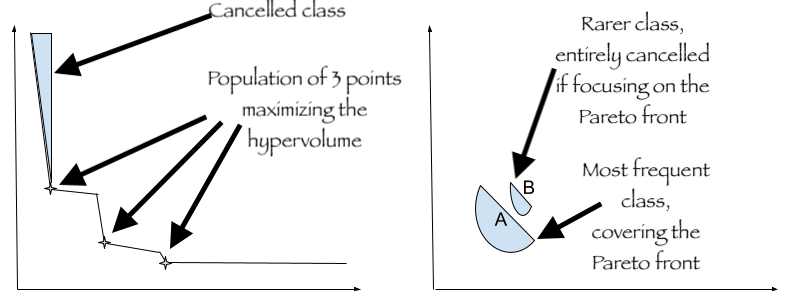}\vspace{-2mm}\caption{\label{moo}Bi-objective minimization, cases in which Pareto-dominance will be detrimental to diversity. Left: artificial counter-example showing that maximizing a numerical diversity criterion (the hypervolume) over the Pareto front might not provide diverse solutions. Here, we see a Pareto-front and the hypervolume-best approximation by 3 points. Dots: the 3 individuals maximizing the hypervolume. Gray areas: examples of classes that completely disappear if we consider those dots (as they maximize the hypervolume) rather than a random sampling of the Pareto front. Right: other counter-example. Class A is assumed to be much bigger than class B, and to have, therefore, better scores for both criteria: this is because, as discussed in the text, bigger classes typically have better scores (see Section~\ref{mcrare}). While local optimization from points in B will provide points in B, a global optimization based on Pareto fronts will provide only points in A: class B is not represented.}
\end{figure}

However, not all methods are equal: we would like to have as much diversity as possible for a given fixed number of proposals. Also, Fig.~\ref{moo} shows that it is not obvious that this will work: though this might not be intuitive, one can design counter-examples in which focusing on the Pareto-front and even more on a few key elements representing the Pareto front can actually decrease the diversity, compared to generating just one image at a time, because the Pareto frontier might be entirely covered by a single class (in particular the biggest class, for which values are usually greater in machine learning models, as explained in Section~\ref{deuxun}).
The simplest, and maybe most robust solution is to run multiple independent (randomized) runs: if the probability $P(g\in C)$ of generating a point in $C$ is  low, then the probability $1-(1-P(g\in C))^k$ of having at least one of $k$ generated image inside $C$ is greater: $1-(1-P(g\in C))^k\geq P(g\in C)$ (strict if $P(g\in C) \not\in\{0,1\}$). If the user needs an image of class $C$, generating $k$ images is more likely to have at least one in $C$ unless the original probability is $0$ or $1$.

The question is now how to do better than this baseline. We consider the following ideas:
\begin{itemize}
	\item the $k$ runs are not using the same weights: e.g., we use random weights in the optimization runs, and they are randomly drawn at each run.
	\item we run a MOO algorithm which tries to maximize some quantity, e.g., the hypervolume of the obtained solutions, or their diversity in the loss space, or the coverage in the domain space.
\end{itemize}
Consistent with the credo of the present paper (not using target classes in the algorithm), these algorithms are independent of the target classes.

\subsection{Stratification by rejection is rarely detrimental}\label{cool}
The reweighting method in Section~\ref{rw} works in the sense that, by design, when we use it, we switch back to the exact probabilities for each stratum, i.e., $p'=p$. This implies that, unless a target class has entirely disappeared in the model, reweighting using strata based on the target classes recovers the frequencies of all target classes. However, the point of the present paper is to fix frequencies of unknown target classes. 
So, now, consider a target class $C$, which is not necessarily one of the strata. If $C$ is one of the $D_i$ (or a union of them) then, as discussed above, the stratification leads to $p(g\in C)=p(\omega\in C)$: let us see if we can find a more general case in which $P(g\in C)=P(\omega\in C)$.

The Diversity Loss (DL) measure we are using (Section~\ref{dl}) for estimating the DL of a model $g$ compared to a random variable $w$ is based on aggregating measures of DL for several classes: the global diversity loss is $\Delta:=1-\inf_{f_i>0} f'_i/f_i$  where $f_i$ is the target frequency for class $i$ and $f'_i$ is the observed frequency.

\begin{eqnarray*}
\Delta&=&\max_{C; P(w\in C)>0} \left( 1-\frac{P(g\in C)}{P(w\in C)}\right)\\
&=&\max_{C; P(w\in C)>0} \frac1{P(w\in C)}\left( {P(w\in C)}-{P(g\in C)}\right)\\
&=&\max_{C; P(w\in C)>0} \frac1{P(w\in C)}\left( pq-pq'\right)
\end{eqnarray*}
where:
\begin{itemize}
    \item $q_j$ is the probability of class $C$ in stratum $D_j$ for the original random variable $w$ i.e. $q_j=P(w\in C| w\in D_j)$;
    \item $q'_j$ is the counterpart for the model $g$ i.e. $q'_j=P(g\in C|g\in D_j)$.
\end{itemize}

The reweighting increases the DL for class $C$ if 
$pq-pq'>pq-p'q'$ (where $pq$ is short for $\sum_j p_jq_j$). This is equivalent to $q'(p'-p)>0$ and $p(q-q')>0$. This means that reweighting is detrimental for this measure if (i) $p(q-q')>0$ and (ii) $q'(p'-p)>0$ occur simultaneously: (i) means that $q-q'$ is overall positive on average for the frequencies $p$ (i.e., $g$ tends to underestimate class $C$), which is precisely the case of interest: this means that $g$ is not doing well on $C$. And (ii) $q'(p'-p) >0$: this implies that we tend to overestimate classes in which $C$ has a low probability, which contradicts the general assumption ``diversity loss usually occurs for rarer classes'' in Section~\ref{genmod}. Therefore, it seems unlikely that reweighting can worsen diversity loss, at least for this measure.

\section{Experimental results}\label{cinq}

\subsection{Framework}
We compare our methods in different contexts.
Each context $(g,b)$ is defined by a generative model $g$ to be compared to a baseline $b$ (dataset or model). 
We check if $g$ has a diversity loss, comparatively to $b$. We have 18 contexts, as described below. 
The baseline $b$ is a dataset or a PGAN~\cite{pgan} trained on it (i.e., two possibilities here), and we try to fix the diversity loss when applying EvolGan~\cite{roziere2020evolgan} with budget 10, 20, 40 (3 possibilities) and algorithm DOPO~\cite{camilleinspir}, CMA~\cite{HAN} or random search (3 possibilities): $g$ can be any of these 9 combinations, and we consider the diversity loss compared to one of the two different possible $b$, hence 18 contexts (Table~\ref{tabtrois1}).
Different contexts have different diversity losses: typically, CMA or RandomSearch lead to more diversity loss than DOPO.

We have checked that (naively) optimizing technical quality is detrimental to fairness (Appendix~\ref{optifair}).  We show (Section~\ref{unrelated}) that applying reweighting according to target classes is unsurprisingly more effective than reweighting according to unrelated strata, but the latter methodology still does mitigate fairness issues. 
Then Section~\ref{moosec} compares various forms of user-assisted optimization for tackling fairness issues.%

\subsection{Reweighting mitigates fairness issues}\label{unrelated}

\subsubsection{Classes A, B, C, D}
\begin{table}
\scriptsize\centering
\begin{tabular}{|c|c|c|c|c|c|}
\hline
	Baseline & Model    & $M$        & Diversity      & Percentage & Percentage \\
	 &         &            &   loss &  of DL & of DL \\
	 &           &                &    before      & remaining & remaining\\
	& & & reweight & with $d=2$ & with $d=4$\\
 \hline
 PGAN &  EG-CMA-10 &  3 &  0.442 & 53.266 & 42.257 \\
 PGAN &  EG-CMA-20 &  3 &  0.513 & 49.901 &     32.176  \\
 PGAN &  EG-CMA-40 &  3 &  0.663 & 83.654 &     40.683 \\
\hline
 PGAN &  EG-D(1+1)-10 &  3 &  0.080 & 74.254 &   16.168     \\
 PGAN &  EG-D(1+1)-20 &  3 &  0.070 & 72.913 &   30.008     \\
 PGAN &  EG-D(1+1)-40 &  3 &  0.115 & 25.079 &   33.147     \\
\hline
 dataset &  EG-RandomSearch-0 &  3 &  0.314 & 31.699 &   25.398 \\
 dataset &  EG-RandomSearch-10 &  3 &  0.563 & 28.083 &  33.860  \\
 dataset &  EG-RandomSearch-20 &  3 &  0.644 & 33.280 &  40.709 \\
 dataset &  EG-RandomSearch-40 &  3 &  0.738 & 65.564 &  63.747 \\
\hline
 dataset &  EG-CMA-0 &  3 &  0.343 & 40.314 &         16.914                  \\
 dataset &  EG-CMA-10 &  3 &  0.561 & 32.505 &        6.927                    \\
 dataset &  EG-CMA-20 &  3 &  0.617 & 27.205 &        29.584                   \\
 dataset &  EG-CMA-40 &  3 &  0.735 & 47.673 &        33.604                   \\
\hline
 dataset &  EG-D(1+1)-0 &  3 &  0.312 & 40.628 &        10.630                   \\
 dataset &  EG-D(1+1)-10 &  3 &  0.339 & 32.440 &       28.977                    \\
 dataset &  EG-D(1+1)-20 &  3 &  0.347 & 95.618 &       11.370                    \\
 dataset &  EG-D(1+1)-40 &  3 &  0.350 & 32.822 &       16.938                    \\
 \hline
 \end{tabular}
\caption{\label{fy}\label{tabtrois2}Impact of reweighting with related variables on the diversity loss for classes A, B, C, D: we see that the original diversity loss is significant (4th column) and reduced a lot if we use 4 variables for reweighting (6th column). Even 2 variables contribute quite well to a significant reduction of diversity loss (5th column). Dataset: faces generated by StyleGAN2. Strata used for reweighting: logits of the output layer of R discretized with $M=3$ and $d=2$ (5th column) or $d=4$ (6th column).}\label{tabquatre}
\end{table}
Table~\ref{fy} presents the diversity loss and the fixed diversity loss when using reweighting.
We use 2 or 4 variables correlated (though not equal) to the target attribute, namely the discretized predicted probabilities of the 4 modalities of the target class. As variables are correlated to the target problem, results are excellent.

\begin{figure}\centering
	\includegraphics[width=.5\textwidth,trim=0 0 0 40,clip]{{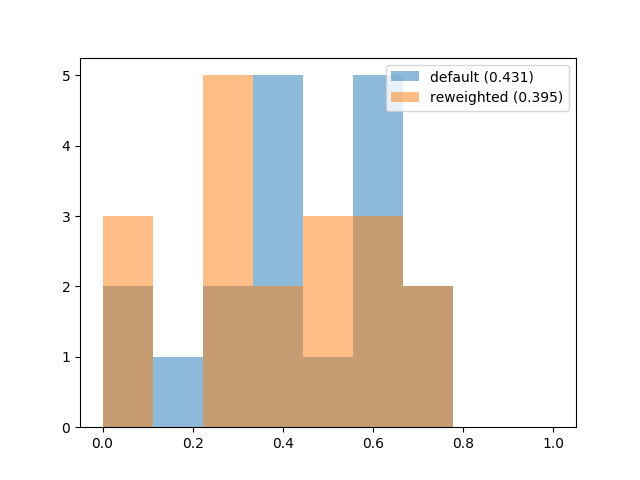}}\vspace{-6mm}
	\caption{\label{hist}Hard case with unrelated reweighting variables: histogram of diversity losses for (A,B,C,D) using reweighting based on strata of R', over each of 18 contexts (see text). The method is slightly beneficial; the average moves from 0.431 to 0.395. We use the best method in Table~\ref{discretization}, rerun from scratch for mitigating the hyperparameter selection bias, getting the same 0.395 value. Dataset, strata, as in Table~\ref{discretization}. X-axis: DL. Y-axis: number of contexts (out of 18) with DL falling in the given DL bin.}
\end{figure}

\begin{table}\centering
\scriptsize\centering
\begin{tabular}{|c|c|c|c|}
\hline
Number of & Discretization & DL before   & DL after\\
vars $d$ & $M$ & reweighting& reweighting \\
\hline
 1  & 2 & \multirow{4}{*}{0.431} & 0.421 \\
 1  & 3 & & 0.428 \\
 1  & 5 & &  0.435  \\
 1  & 8 &  & 0.430  \\
\hline
 2  & 3 & \multirow{3}{*}{0.431} &  0.403 \\
 2  & 5 &  & 0.431 \\
 2  & 8 &  & 0.433 \\
\hline
  4  & 2 & \multirow{4}{*}{0.431} &  0.414 \\
 4  & 3 &  & 0.403 \\
 4  & 5 &  & 0.428 \\
 4  & 8 &  & 0.427 \\
\hline
 10  & 3 & \multirow{2}{*}{0.431} & 0.395  \\
 10  & 5 &  & 0.423 \\ 
\hline
 20  & 2 & \multirow{4}{*}{0.431} & 0.419 \\
 20  & 3 &  & 0.401 \\
 20  & 5 &  & 0.432  \\
 20  & 8 &  & 0.428 \\
\hline
 80  & 2 & \multirow{2}{*}{0.431} & 0.419 \\
 80  & 8 &  & 0.428 \\
 \hline
\end{tabular}
	\caption{\label{discretization} Diversity loss for (A, B, C, D) after reweighting, in our hardest context (variables very uncorrelated to the target variable, namely E'). We observe that in most cases, the reweighting is still beneficial compared to 0.431 originally, though this difficult case does not lead to drastic improvements. Dataset: faces generated by StyleGan2. Strata: discretization of E' with $d\in\{1,2,4,10,20,80\}$ and $M\in \{2,3,5,8\}$.}\label{tabcinq}
\end{table}
We now switch to a more challenging case. Table~\ref{discretization} compares various discretizations in the difficult context of reweighting variables unrelated to the target variables. E.g. (80,8) means that we use $d=80$ variables and split each of them in $M=8$ bins. We got the best results with 10 variables discretized in 3.
There are four target classes for faces unrelated to emotions. The variables are the final layer %
of an emotion recognition network.
Still, in that difficult case, Fig.~\ref{hist} shows how diversity losses are moved in the right direction by the reweighting -- not much, but beneficial, and most importantly, not detrimental.  

\subsubsection{Class E: confirming results for reweighting with unrelated variables}\label{classe}

\begin{table}\scriptsize\centering\setlength{\tabcolsep}{4.7pt}
\begin{tabular}{|c|c|c|c|c|c|}
\hline
Source & Target &           $d$ &  $M$  & DL & Remaining DL(\%)\\
\hline
PGAN &   EG-CMA 10 & 1 & 8 & 0.675 & 99.832 \\
PGAN &   EG-CMA 20 & 1 & 8 & 0.778 & 100.444 \\
PGAN &   EG-CMA 40 & 1 & 8 & 0.872 & 100.923 \\
\hline
PGAN &   EG-D(1+1) 10 & 1 & 8 & 0.108 & 103.808 \\
PGAN &   EG-D(1+1) 20 & 1 & 8 & 0.204 & 76.961 \\
PGAN &   EG-D(1+1) 40 & 1 & 8 & 0.333 & 92.000 \\
\hline
PGAN &   EG-RandomSearch 10 & 1 & 8 & 0.675 & 88.797 \\
PGAN &   EG-RandomSearch 20 & 1 & 8 & 0.785 & 100.171 \\
PGAN &   EG-RandomSearch 40 & 1 & 8 & 0.876 & 98.348 \\
\hline
\hline
PGAN &  EG-CMA-10 & 2 & 8 & 0.675 & 96.660 \\
PGAN &  EG-CMA-20 & 2 & 8 & 0.778 & 95.140 \\
PGAN &  EG-CMA-40 & 2 & 8 & 0.872 & 89.906 \\
\hline
PGAN &  EG-D(1+1)-10 & 2 & 8 & 0.108 & 103.808 \\
PGAN &  EG-D(1+1)-20 & 2 & 8 & 0.204 & 89.274 \\
PGAN &  EG-D(1+1)-40 & 2 & 8 & 0.333 & 98.751 \\
\hline
PGAN &  EG-RandomSearch-10 & 2 & 8 & 0.675 & 87.870 \\
PGAN &  EG-RandomSearch-20 & 2 & 8 & 0.785 & 91.072 \\
PGAN &  EG-RandomSearch-40 & 2 & 8 & 0.876 & 97.075 \\
\hline
\hline
PGAN &  EG-CMA-10 & 3 & 8 & 0.675 & 95.538 \\
PGAN &  EG-CMA-20 & 3 & 8 & 0.778 & 95.218 \\
PGAN &  EG-CMA-40 & 3 & 8 & 0.872 & 89.552 \\
\hline
PGAN  &  EG-D(1+1)-10 & 3 & 8 & 0.108 & 90.259 \\
PGAN  &  EG-D(1+1)-20 & 3 & 8 & 0.204 & 81.139 \\
PGAN  &  EG-D(1+1)-40 & 3 & 8 & 0.333 & 84.623 \\
\hline
PGAN &  EG-RandomSearch-10 & 3 & 8 & 0.675 & 86.845 \\
PGAN &  EG-RandomSearch-20 & 3 & 8 & 0.785 & 87.502 \\
PGAN &  EG-RandomSearch-40 & 3 & 8 & 0.876 & 95.705 \\
\hline
\hline
PGAN &  EG-CMA-10 & 4 & 8 & 0.675 & 97.587 \\
PGAN &  EG-CMA-20 & 4 & 8 & 0.778 & 96.505 \\
PGAN &  EG-CMA-40 & 4 & 8 & 0.872 & 90.098 \\
\hline
PGAN  &  EG-D(1+1)-10 & 4 & 8 & 0.108 & 70.592 \\
PGAN  &  EG-D(1+1)-20 & 4 & 8 & 0.204 & 94.112 \\
PGAN  &  EG-D(1+1)-40 & 4 & 8 & 0.333 & 88.999 \\
\hline
PGAN &  EG-RandomSearch-10 & 4 & 8 & 0.675 & 87.333 \\
PGAN &  EG-RandomSearch-20 & 4 & 8 & 0.785 & 88.270 \\
PGAN &  EG-RandomSearch-40 & 4 & 8 & 0.876 & 96.438 \\
 \hline
 \end{tabular}
 \caption{\label{cE}Impact of reweighting on diversity loss for class E when using classes R as auxiliary variable. We see that adding variables almost always improves results, and cases in which reweighting is detrimental are rare. Dataset: faces generated by StyleGan2. Sensitive variables for which DL is computed: emotions. Strata:  IQA values provided by R', i.e., logits of R, with discretization with $d\in\{1,2,3,4\}$ variables and $M=8$ equally likely bins per variable. Observation: increasing $d$ reduces the DL after reweighting.}\label{tabsix}
 \end{table}
Table~\ref{cE} presents the impact of reweighting using the probabilities of class A, B, C and D (discretized) on the diversity loss of class E. (ABCD) and E are unrelated, so this is unsupervised fairness improvement.

\subsection{Multi-objective optimization: only some forms of MOO mitigate fairness issues}\label{moosec}
MOO typically has two phases:
\begin{itemize}
	\item optimization run, building a possibly large Pareto front;
	\item selection of a reduced Pareto front for presentation to the user.
\end{itemize}
This does not cover all MOO methods. The second stage is not always present, as some tools are equipped with a mechanism for navigating the Pareto front. Also, sometimes the first stage includes inputs from the human. We will nonetheless consider the framework above in the present paper. As mentioned before, a simple solution for MOO is to do multiple simple runs (MSR): just run the algorithm several times, and consider the several outputs. We consider other methods, namely maximizing the hypervolume for phase 1 and using various techniques (IGD, EPS, RANDOM, see Appendix~\ref{subpareto}) for constructing a subset. 

\begin{table}\centering
\scriptsize\centering
\begin{tabular}{|c|c|c|}
\hline
Algorithm & Selector & Percentage \\
\hline
\multicolumn{3}{|c|}{9 single-objective runs}\\
\hline
	NGOpt 9 & domain-covering & {\bf{33}} \\
	NGOpt 9 & eps & {\bf{33}} \\
	NGOpt 9 & loss-covering & {\bf{33}} \\
	NGOpt 9 & msr & {\bf{33}} \\
\hline
\multicolumn{3}{|c|}{CMA}\\
\hline
CMA & domain-covering & 33 \\
CMA & eps & 33 \\
CMA & loss-covering & 44 \\
CMA & {\bf{msr}} & {\bf{66}} \\
\hline
\multicolumn{3}{|c|}{Portfolio Discrete-$(1+1)$}\\
\hline
PortfolioDiscrete$(1+1)$ & msr & 16 \\
PortfolioDiscrete$(1+1)$ & eps & 33 \\
PortfolioDiscrete$(1+1)$ & loss-covering & 33 \\
PortfolioDiscrete$(1+1)$ & {\bf{domain-covering}} & {\bf{83}} \\
\hline
\multicolumn{3}{|c|}{Differential Evolution}\\
\hline
DE & loss-covering & 16 \\
DE & eps & 16 \\ 
DE & domain-covering & 33 \\ 
DE & {\bf{msr}} & {\bf{55}} \\ 
\hline
\multicolumn{3}{|c|}{Random Search}\\
\hline
RandomSearch & loss-covering & 0 \\
RandomSearch & msr & 33 \\
RandomSearch & eps & 50  \\
RandomSearch & {\bf{domain-covering}} & {\bf{66}} \\
\hline
\end{tabular}
\caption{\label{mooinspir}\label{six}Multi-objective inspirational generation: the target is the face of a black person, originally very pixelized; the goal is to approximate it with PytorchGanZoo. We consider with which probability PytorchGanZoo generates at least one face of the correct ethnicity. Each algorithm generates nine faces. 
The best selector consists of picking up the nine outcomes of nine single runs (MSR: multiple single runs) or using domain covering, i.e., never using a Pareto-based measure. In conclusion, multi-objective optimization does work for generating diversity. However, we should not use Pareto-dominance and focus on multiple outcomes of random single-objective runs or diversity in the domain (``domain-covering'' method), because fitness-based measures are too biased for being used for diversity.}\label{tabhuit}
\end{table}

\begin{table}\centering\centering
\scriptsize
\begin{tabular}{|c|c|c|}
\hline
Algorithm & Selector & Percentage \\
\hline
\multicolumn{3}{|c|}{9 single-objective runs}\\
\hline
NGOpt 9 & {\bf{domain-covering}} & {\bf{22}} \\
NGOpt 9 & eps & 0 \\
NGOpt 9 & loss-covering & 5 \\
NGOpt 9 & msr & 11 \\
\hline
\multicolumn{3}{|c|}{CMA}\\
\hline
CMA & {\bf{domain-covering}} & {\bf{27}} \\
CMA & eps & 11 \\
CMA & loss-covering & 22 \\
CMA & {{msr}} & {{0}} \\
\hline
\multicolumn{3}{|c|}{Portfolio Discrete-$(1+1)$}\\
\hline
PortfolioDiscrete$(1+1)$ & msr & 0 \\
	PortfolioDiscrete$(1+1)$ & eps & {{38}} \\
PortfolioDiscrete$(1+1)$ & loss-covering & 16 \\
PortfolioDiscrete$(1+1)$ & {{domain-covering}} & {{38}} \\
\hline
\multicolumn{3}{|c|}{Differential Evolution}\\
\hline
DE & loss-covering & 16 \\
DE & {\bf{eps}} & {\bf{22}} \\ 
DE & domain-covering & 5 \\ 
DE & {{msr}} & {{11}} \\ 
\hline
\multicolumn{3}{|c|}{Random Search}\\
\hline
RandomSearch & loss-covering & 5 \\
RandomSearch & msr & 0 \\
RandomSearch & eps & 11  \\
RandomSearch & {\bf{domain-covering}} & {\bf{33}} \\
\hline
\hline
\end{tabular}
\caption{\label{mooinspir2}\label{sept}Counterpart of Table~\ref{mooinspir} for female Asian target. As in Table~\ref{mooinspir}, domain-covering performs best.}\label{tabneuf}
\end{table}
Tables~\ref{mooinspir} (target class is black) and~\ref{mooinspir2} (target class is female Asian) show that the best results concerning maximum diversity are obtained by domain-covering or by MSR, and not by MOO approaches focusing on diversity over the Pareto front. The effective diversity measures are not based on Pareto-dominance. The best results are obtained either by pure MSR, using multiple runs and keeping all results, or by domain-covering, i.e., creating a subset using diversity in the image domain.
This result is not so intuitive, so we ran additional experiments to check if Pareto-dominance can be detrimental to diversity.
\begin{table}
\scriptsize \centering
\begin{tabular}{|c|c|c|c|c|c|c|}
\hline
Original & EG40 & PF &  Subset & $d,M$ & Diversity& Uncancelled \\
model & variant & size &  &   & loss & loss (\%)\\
 \hline
PGAN&  EG-RandomSearch & 16 &  COV & 5,2  & 0.726 & 111.333 \\
PGAN&  EG-CMA & 16 &  COV & 5,2 & 0.977 & 89.285 \\
PGAN&  EG-D(1+1) & 16 &  COV & 5,2 & {\bf{0.707}} & 116.297 \\
 \hline
PGAN&  EG-RandomSearch & 16 &  IGD & 5,2 & {\bf{0.72}} & 112.165 \\
PGAN&  EG-CMA & 16 &  IGD & 5,2 & 0.973 & 90.008 \\
PGAN&  EG-D(1+1) & 16 &  IGD & 5,2 & 0.730 & 112.409 \\
 \hline
PGAN&  EG-RandomSearch & 16 &  Random & 5,2 & {\bf{0.697}} & 118.724 \\
PGAN&  EG-CMA & 16 &  Random & 5,2 & 0.969 & 89.622 \\
PGAN&  EG-D(1+1) & 16 &  Random & 5,2 & 0.726 & 115.778 \\
 \hline
PGAN&  EG-RandomSearch & 16 &  EPS & 5,2 & 0.738 & 107.916 \\
PGAN&  EG-CMA & 16 &  EPS & 5,2 & 0.977 & 88.095 \\
PGAN&  EG-D(1+1) & 16 &  EPS & 5,2 & {\bf{0.709}} & 116.377 \\
 \hline
\end{tabular}
	\caption{\label{moosolve}\label{huit}Column 6 shows the DL when moving from the original (column 1) to the improved version (column 2), and column 7 presents the part of this DL which is not solved by applying MOO for generating 16 points. There is a strong computational budget (10000) and a large generated set (16 points) in the present context. We consider that the result is ok if at least one of those 16 generations is of the expected class. Column 7 is frequently above 100\%, i.e., results are {\bf{worse}} than in the single-objective case generating only one image: this shows that even with favorable conditions, MOO based on Pareto-dominance can be detrimental. Only MSR (running several times and gathering the results) or domain-covering (i.e., good diversity for a side measure in the domain) provide stable improvements in the user-assisted context (Tables~\ref{six} \&~\ref{sept}). Dataset: CelebaHQ (see \url{https://github.com/tkarras/progressive_growing_of_gans}). Model: PytorchGanZoo. Method: described in Section~\ref{troistrois}. Sensitive variables on which DL is measured: ethnicity.}
\end{table}
{\bf{We conclude that Pareto-based MOO can be detrimental to diversity even with a large budget and 16 generations instead of 1.}}
This is shown by Table~\ref{moosolve}: we do an additional experiment based on Pytorch-Gan-ZOO and variants. We use both single-objective optimization (EvolGan with budget 10000) and our MOO counterpart. We get a single image per run for single-objective optimization, and we can estimate DL as usual. We use MOO, with three objectives linearly combined in the single-objective case: minimizing the squared of the injected latent variables, maximizing the IQA score, and maximizing the discriminator score. We use a large budget and many generated individuals so that problems can not be attributed to the parametrization. We consider that the ``frequency'' of a class is the frequency at which at least one of the outputs contains that class (see Alg.~\ref{algmultiple}). 
We see that MOO by classical Pareto-dominance is not always solving diversity issues. It works only when the method has over-optimized and completely destroyed diversity (Table~\ref{moosolve}: results are $<100\%$ in the last column only if the diversity loss is $>95\%$). Whereas diversity in the domain (domain-covering) or simple multiplication of runs (as in MSR) works in many cases, optimization with Pareto-dominance can fail.
We conclude that counter-examples as in Fig.~\ref{moo} are not an exception but the standard behavior of Pareto-dominance: due to different scales of quality depending on the frequency of classes, we can not reliably use Pareto-dominance for selecting samples.
MSR is the only method that did not have counter-examples. MOO methods based on Pareto fronts were ok only when the method for extracting representative images was based on domain-covering, i.e., unsupervised correction.

\section{Conclusion}
{\bf{Quality improvement degrades diversity:}} We checked that improving the visual quality degrades diversity when biasing latent variables through IQA methods. The biasing effect is consistent with known facts.

To mitigate this issue, we propose two methods. The first (Alg.~\ref{algrw}) is a drop-in improvement of a generative model: it can be applied as soon as we have some auxiliary features that we can use for defining strata. The second one is user-assisted (Alg.~\ref{algmultiple}) and can use MOO (either with Pareto-dominance for selecting a subset or with diversity preservation for some features in the domain) or MSR.

\noindent
{\bf{Reweighting by related auxiliary variables:}} Unsurprisingly, reweighting by auxiliary variables close to the target classes is very effective at reducing the diversity loss. We cancel the diversity loss when reweighting using the same target class. This incurs a computational cost and does not solve quality inside each class, but we recover target frequencies.

\noindent
{\bf{Reweighting by unrelated auxiliary variables:}} A good finding is that we never degrade performance by applying reweighting, even when using unrelated variables. There are good reasons for this (Section~\ref{cool}). We recommend reweighting by as many variables as possible (at least as long as there is data enough for computing statistics with enough precision). However, we acknowledge that this has a computational cost.

\noindent
{\bf{Using MOO, also without knowing categories:}}
The idea of using MOO for generating diversity is intuitively appealing. However, only MSR (running several single objective problems) or domain-covering turned out to be effective. Methods based on Pareto-dominance can be detrimental. Phenomena, as described in Fig.~\ref{moo}, are not an exception, but the rule.

\subsection*{Side remarks \& caveats}

{\bf{Combination with supervised fairness:}} we considered purely unsupervised fairness, but we could do the same in combination with given sensitive variables: after a first correction for given sensitive variables, we can add a correction with respect to some unrelated generic strata.

\noindent
{\bf{Impact of the optimization method:}} Tables~\ref{tabtrois1},~\ref{tabquatre} and~\ref{huit} show that CMA leads to more diversity loss compared to random search or PortfolioDiscrete(1+1). This is reasonable as the prior distribution is ignored by CMA, whereas it impacts every  other tested methods:
\begin{itemize}
    \item Random search uses the prior distribution at each step for choosing a point;
    \item Discrete $(1+1)$ algorithms use the marginal of the probability distribution for each modified variable.
\end{itemize}
We presented results for reweighting with statistics based on large datasets, so that there was no problem for precisely estimating $p_i/p_j$ as needed: with small datasets, precision might be an issue.

\appendix

\section*{Appendix}

\section{Subsampling the Pareto front}\label{subpareto}

To extract $1\leq m\leq n$ points from an approximate Pareto set $\{x_1,\dots,x_n\}$, a range of approaches can be used:
\begin{itemize}
    \item Random subset: just pick up $m$ of the $x_i$, {uniformly at random and} without replacement.
    \item HV: pick up $\{x_{j_1},\dots,x_{j_m}\}$ such that their Hypervolume $C_h$ is maximal. 
    \item Loss-covering, also known as IGD (inverted generational distance,~\cite{sato2004igd}): pick up $\{x_{j_1},\dots,x_{j_m}\}$ such that \\$C_{l}=\sum_{i=1}^n \inf_{j\leq m} ||F(x_i)-F(x_{i_j})||^2$ is minimal, where $F(x)=(f_1(x),\dots,f_N(x))$.
    \item COV (covering the Pareto-front): pick up $\{x_{j_1},\dots,x_{j_m}\}$ such that $C_{d}=\sum_{i=1}^n \inf_{j\leq m} ||x_i-x_{i_j}||^2$ is minimal.
    \item Additive epsilon approximation (EPS,~\cite{Papadimitriou2000eps}): pick up $\{x_{j_1},\dots,x_{j_m}\}$ such that $C_{e}=\max_{i=1}^n \inf_{j\leq m} ||F(x_i)-F(x_{i_j})||_\infty$ is minimal, where $F(x)=(f_1(x),\dots,f_N(x))$. %
\end{itemize}
In domain-covering, we do the same as COV, but over all generated points and not only the Pareto-front.

\section{(Naively) optimizing $\to$ less diversity}\label{optifair}

We train a PGAN~\cite{pgan} and then improve it using IQA as in \cite{roziere2020evolgan}: PGAN $\to$ EG10 $\to$ EG20 $\to$ EG40 (each ``$\to$'' being an improvement in terms of image quality by refining the latent variables using the image quality assessment tool as a criterion\cite{roziere2020evolgan}).
As noted in \cite{roziere2020evolgan}, the quality improvement in EvolGAN is related to some diversity losses: for horses, we get rid of bugs such as horses with 3 heads, which is in some sense a sort of diversity loss. Unfortunately, this also reduces diversity in the sense that relevant rare classes become rarer (Table~\ref{optloss}):
there is a diversity loss from the dataset to the PGAN, and this diversity loss is increased when we increase the budget of the GAN improvement by EvolGan.

\begin{table}[t]\centering
\scriptsize\centering
 \begin{tabular}{|c|c|c|}
\hline
	 EG  & Diversity  & Remaining \\
	 variant & loss & diversity loss (\%)\\
\hline
EG-CMA-10 & 0.675 & 97.587 \\
EG-CMA-20 & 0.778 & 96.505 \\
EG-CMA-40 & 0.872 & 90.098 \\
\hline
EG-D(1+1)-10 & 0.108 & 70.592 \\
EG-D(1+1)-20 & 0.204 & 94.112 \\
EG-D(1+1)-40 &  0.333 & 88.999 \\
\hline
EG-RandomSearch-10 &  0.675 & 87.333 \\
EG-RandomSearch-20 &  0.785 & 88.270 \\
EG-RandomSearch-40 &  0.876 & 96.438 \\
\hline
 \end{tabular}
	\caption{\label{optloss}Diversity loss for class F (i.e., low aesthetics value according to AvA) for EG compared to PytorchGanZoo (EG is an improvement of PytorchGanZoo using K512 as an IQA for biasing the latent vaariables). The diversity loss depends on how strongly we improve the GAN using EvolGan (more budget = more improvement in terms of quality measured by K512). We also show (third column) how much the diversity loss is preserved in spite of reweighting w.r.t. $E$: numbers $<100\%$ show that a part of the diversity loss is repaired. No number is greater than $100\%$: our method is never detrimental. }\label{tabtrois1}
\end{table}

{\bf{Further appendices can be found in the supplementary material.}}
\FloatBarrier
\bibliographystyle{abbrv}
\bibliography{references,icml2021,camille,biblio}
\clearpage

\section*{Supplementary material for: Fairness in Generative Modeling: do it Unsupervised!}

\emph{Disclaimer.} 
We understand 
that neither the reviewers, the track chairs, nor the editors are required to look at the supplementary material. We also understand that they are requested to base their reviews and decisions solely on the main PDF manuscript.

\subsection*{(SUP1) Human raters}
We use human raters for the two applications in Tables~\ref{mooinspir} and~\ref{mooinspir2}. Our raters are volunteers, without any time limit.
A double-blind graphical user interface presents images.
For labeling with ethnicity, we use a binary question.

\subsection*{(SUP2) Reweighting with respect to four $VF$ binarized variables for a specific target}

\begin{table}[!h]\begin{scriptsize}
\begin{tabular}{|c|c|c|c|c|}
\hline
& & StyleGan2 & EG & Reweigh -\\
& &  &  & EG\\\hline
	\multicolumn{5}{|c|}{Selection rate in EG: 5.1\%}\\
\hline
100/1979 &corr. & .0480  & .0297  & .0258  \\
100/1979 & random & .0480  & .0297  & .0309  \\
\hline
	\multicolumn{5}{|c|}{Selection rate in EG: 7.6\%}\\
\hline
150/1979 &corr. & .0480  & .0271  & .0219  \\
150/1979 & random & .0480  & .0271  & .0292  \\
\hline
	\multicolumn{5}{|c|}{Selection rate in EG: 10.1\%}\\
\hline
200/1979 &corr. & .0480  & .0247  & .0207  \\
200/1979 & random & .0480  & .0247  & .0269  \\
\hline
	\multicolumn{5}{|c|}{Selection rate in EG: 12.6\%}\\
\hline
250/1979 &corr. & .0480  & .0273  & .0279  \\
250/1979 & random & .0480  & .0273  & .0277  \\
\hline
	\multicolumn{5}{|c|}{Selection rate in EG: 15.2\%}\\
\hline
300/1979 &corr. & .0480  & .0239  & .0245  \\
300/1979 & random & .0480  & .0239  & .0234  \\
\hline
	\multicolumn{5}{|c|}{Selection rate in EG: 17.7\%}\\
\hline
350/1979 &corr. & .0480  & .0285  & .0333  \\
350/1979 & random & .0480  & .0285  & .0300  \\
\hline
	\multicolumn{5}{|c|}{Selection rate in EG: 20.2\%}\\
\hline
400/1979 &corr. & .0480  & .0353  & .0350  \\
400/1979 & random & .0480  & .0353  & .0374  \\
\hline
	\multicolumn{5}{|c|}{Selection rate in EG: 22.8\%}\\
\hline
450/1979 &corr. & .0480  & .0358  & .0370  \\
450/1979 & random & .0480  & .0358  & .0390  \\
\hline
	\multicolumn{5}{|c|}{Selection rate in EG: 25.3\%}\\
\hline
500/1979 &corr. & .0480  & .0344  & .0354  \\
500/1979 & random & .0480  & .0344  & .0360  \\
\hline
	\multicolumn{5}{|c|}{Selection rate in EG: 27.8\%}\\
\hline
550/1979 &corr. & .0480  & .0344  & .0364  \\
550/1979 & random & .0480  & .0344  & .0363  \\
\hline
	\multicolumn{5}{|c|}{Selection rate in EG: 25.3\%}\\
\hline
600/1979 &corr. & .0480  & .0331  & .0340  \\
600/1979 & random & .0480  & .0331  & .0356  \\
\hline
\end{tabular}\end{scriptsize}
	\caption{\label{stylegan2}Dataset: faces generated by StyleGan2. The frequency of black people in the different versions, depending on which strata are used for applying the reweighting method of Section~\ref{troisdeux}. Random: four variables randomly picked up among the 128 binary variables built from VGG-Faces. Correlated: same VGG-Faces, but we use the most correlated ones.}%
\end{table}

Table~\ref{stylegan2} presents results of different methods in terms of the frequency of black people. In most cases, the frequency of black people decreased from the original 4.8\% when applying EvolGan, but increased when applying reweighting. We note exceptions: whereas randomly chosen variables were always beneficial, very correlated variables failed in the most difficult cases.

\end{document}